\documentclass[sn-mathphys,Numbered]{sn-jnl}

\usepackage{graphicx}%
\usepackage{multirow}%
\usepackage{amsmath,amssymb,amsfonts}%
\usepackage{amsthm}%
\usepackage{mathrsfs}%
\usepackage[title]{appendix}%
\usepackage{xcolor}%
\usepackage{textcomp}%
\usepackage{manyfoot}%
\usepackage{booktabs}%
\usepackage{algorithm}%
\usepackage{algorithmicx}%
\usepackage{algpseudocode}%
\usepackage{listings}

\usepackage{acronym}
\usepackage{xurl}
\usepackage{algorithm}
\usepackage{algpseudocode}
\usepackage{amsmath}

\acrodef{SNN}[SNN]{Spiking Neural Network}
\acrodef{STDP}[STDP]{Spike-Timing-Dependent Plasticity}
\acrodef{ANN}[ANN]{Artificial Neural Network}
\acrodef{DNN}[DNN]{Deep Neural Network}
\acrodef{LIF}[LIF]{Leaky Integrate-and-Fire}
\acrodef{ISI}[ISI]{Inter-Spike Interval}
\acrodef{ToF}[ToF]{Time of Flight}

\begin{document}

\title{A Low-Cost Real-Time Spiking System for Obstacle Detection based on Ultrasonic Sensors and Rate Coding}

\author*[1]{\fnm{Alvaro} \sur{Ayuso-Martinez}}\email{aayuso@us.es}
\author[1]{\fnm{Daniel} \sur{Casanueva-Morato}}\email{dcasanueva@us.es}
\author[1]{\fnm{Juan Pedro} \sur{Dominguez-Morales}}\email{jpdominguez@us.es}
\author[1]{\fnm{Angel} \sur{Jimenez-Fernandez}}\email{angel@us.es}
\author[1]{\fnm{Gabriel} \sur{Jimenez-Moreno}}\email{gaji@us.es}
\author[2]{\fnm{Fernando} \sur{Perez-Peña}}\email{fernandoperez.pena@uca.es}

\affil*[1]{\orgdiv{Computer Architecture and Technology}, \orgname{Universidad de Sevilla}, \orgaddress{\street{Av. de la Reina Mercedes, s/n}, \city{Sevilla}, \postcode{41012}, \state{Andalucia}, \country{Spain}}}

\affil[2]{\orgdiv{Automation, Electronics, Architecture and Computer Networks}, \orgname{Universidad de Cádiz}, \orgaddress{\street{Av. Universidad de Cádiz, 10}, \city{Puerto Real}, \postcode{11519}, \state{Andalucia}, \country{Spain}}}

\abstract{Since the advent of mobile robots, obstacle detection has been a topic of great interest. It has also been a subject of study in neuroscience, where flying insects and bats could be considered two of the most interesting cases in terms of vision-based and sound-based mechanisms for obstacle detection, respectively. Currently, many studies focus on vision-based obstacle detection, but not many can be found regarding sound-based obstacle detection. This work focuses on the latter approach, which also makes use of a Spiking Neural Network to exploit the advantages of these architectures and achieve an approach closer to biology. The complete system was tested through a series of experiments that confirm the validity of the spiking architecture for obstacle detection. It is empirically demonstrated that, when the distance between the robot and the obstacle decreases, the output firing rate of the system increases in response as expected, and vice versa. Therefore, there is a direct relation between the two. Furthermore, there is a distance threshold between detectable and undetectable objects which is also empirically measured in this work. An in-depth study on how this system works at low level based on the Inter-Spike Interval concept was performed, which may be useful in the future development of applications based on spiking filters.}

\keywords{Obstacle detection, Ultrasonic sensor, Spiking Neural Network, Spiking filter, Rate coding}

\maketitle


\section{Introduction}
\label{sec:introduction}


Obstacle detection and avoidance has been a topic of interest in the field of robotics since the advent of mobile robots more than fifty years ago \cite{tzafestas2013introduction}. When talking about autonomous navigation, two main common problems arise: firstly, when the aim of the robot is to reach an end point starting from an initial point, it has to find a way to avoid obstacles that may exist on its path in an optimal way, something known as path planning. Secondly, mobile robots always have to deal with the appearance of unexpected obstacles that may cross their path, something that in real applications is essential to guarantee the safety of the robot and, in the case of that robot being a vehicle, also of its passengers. 


The task of detecting obstacles is not an easy one. Its accuracy usually depends on the shape of the obstacle to be detected and, as mentioned in \cite{hu2020survey}, it involves sensor characteristics and known problems, and environmental conditions. In \cite{hu2020survey}, the most commonly used methods for obstacle detection in intelligent ground vehicles are also collected and compared according to relevant characteristics such as detection range, robustness and cost, mentioning the main problems encountered in each of them. 

In this way, sensors most commonly used for this task can be summarized in four types: SONAR, LIDAR, RADAR and cameras. Most vehicles do not use a single type of sensors, but different types of them. This sensory fusion makes it possible to solve or smooth the known problems of each of the sensors used and to exploit new advantages, in exchange for a certain added computational cost. Although there are many examples of this sensory fusion, one of the most interesting is the following: one of the major problems of LIDAR arises when the object to be detected is a translucent object, since there is no reflection of the emitted light similar to that produced with opaque objects. However, this problem does not exist in SONAR because it uses sound waves for sensing. Thus, it could be said that the two can complement each other to form a fairly robust obstacle detection method. In this way, there are many works in which multiple sensors are used for obstacle detection \cite{john2019rvnet, chang2020spatial, hajri2018real, gholami2022real}. 


From a biological point of view, it is obstacle detection and collision avoidance that allow animals to navigate complex environments, which is necessary to perform other vital tasks such as foraging for food or escaping from a predator. Flying insects are the most studied to understand how these functions are performed in biology, due to their high precision and speed in avoiding these obstacles, even at night or in poorly lit environments. Although active sensors are generally used in robotics to carry out these tasks, the reality is that these flying insects perform these tasks using mainly vision \cite{bertrand2015bio}, something that can be extended to mammals and other animals. In \cite{kestur2012emulating}, it is discussed without going into much biological detail how the human visual system is able to focus attention on regions of interest in a visual scene as a function of the objects recognized in that region, which is directly related to the task of obstacle detection and, furthermore, how mammals are able to navigate complex environments. Other works provide greater neuroscientific knowledge about how navigation occurs in mammals \cite{poulter2018neurobiology, wood2021navigating, de2022predictive}.


However, although it might seem that the problem of navigation and avoidance of objects in the environment is already solved thanks to vision, the truth is that this is not always the case. When light conditions are particularly poor, vision is no longer a possibility and animals have to resort to other mechanisms to perform this task. The best known of these consists of emitting high-frequency sounds, also known as ultrasounds, and measuring the time it takes for the echo to return. In this way, an object will be closer to the emitter the shorter this time is. This mechanism, commonly referred as echolocation, is used by many animal species, where bats, dolphins and toothed whales are just some of the most interesting examples. These ultrasounds can also be used to extract extra information; for example, it is known that some types of bats use them to classify insects based on frequency patterns in the echoes \cite{von1990classification}. Some other works study how bats are able to produce ultrasounds and process their echoes \cite{moss2018auditory, kossl1995basilar}.


In recent years, many works have focused on vision to design neuromorphic applications, i.e., bio-inspired applications with the aim of mimicking the biological behavior of animals \cite{mead1990neuromorphic}, in which the obstacle detection task is performed using bio-inspired vision sensors \cite{blum2017neuromorphic, salt2017obstacle, hu2016bio, milde2017obstacle}. Thus, vision seems to be the preferred sense for the design of bio-inspired systems and algorithms for obstacle detection, while sound-based systems usually focus on other tasks such as sound classification and localization or speech recognition \cite{dominguez2018deep, wall2012spiking, wu2020deep}.

However, it is also possible to find some works in the literature that focus on sound to perform this task. The most interesting is \cite{arena2009learning}, in which a \ac{SNN} capable of performing the obstacle detection task in a mobile robot by using two range sensors, which could be ultrasonic sensors, is presented. This paper highlights the implementation of a spiking application that allows the mobile robot to navigate autonomously; however, the obstacle detection task is not performed in a purely spiking manner, since it is performed based on the digital comparison of two values (the current distance and the threshold distance), which triggers the activation of a specific sensory neuron in the network. 



On the other hand, the implementation of a spiking application purely based on \acp{SNN} for obstacle detection is quite interesting to exploit the advantages of this bio-inspired paradigm, which are mainly low power consumption and high real-time capability. Real-time capability is a critical point in the development of robotic applications, since it determines how fast a robot is able to interact with its environment in a deterministic way. Thus, greater real-time capability translates into a greater ability to react to the appearance of unexpected obstacles in a robot's path.

In this work, a purely spike-based obstacle detection system is proposed, studied and implemented. This system focuses on encoding digital information into spiking information and processing it for the development of an obstacle detection application in the field of robotics.

Ultrasonic sensors are used in this case mainly for two reasons: firstly, ultrasonic sensors provide a very cheap and simple alternative for the development of robotic applications; secondly, this type of sensor allows experiments to be carried out with a certain similarity to how echolocation occurs in animals, as explained above, which may be particularly useful for future work to achieve a more bio-inspired approach.

The main contributions of this work include the following:

\begin{itemize}
    \item Development of a purely SNN-based obstacle detection system
    \item Low-level analysis of the implemented \ac{SNN} and the encoded information
    \item The code used in this work is publicly-available in a GitHub repository and has been released under a GPL license\footnote{\url{https://github.com/alvayus/spiking_rtod}}
\end{itemize}

The rest of the paper is structured as follows: in Section~\ref{sec:materials_and_methods} information regarding software and hardware materials used in this work is given; Section~\ref{sec:system} explains how these materials are interconnected in the global system architecture to perform obstacle detection; in Section~\ref{sec:snn_design} the design of the implemented \ac{SNN} is shown and explained; Section~\ref{sec:experiments_and_results} details different experiments carried out to test the performance of the implemented system; in Section~\ref{sec:discussion} the results obtained from experimentation and high-level details are discussed in depth; finally, the conclusions of the work are presented in Section~\ref{sec:conclusions}.
\section{Materials and methods}
\label{sec:materials_and_methods}

This section presents the hardware and software components used for the development of the complete system. The most relevant details of each of them are shown in the following subsections.

\subsection{Robotic platform}
\label{subsec:robotic_platform}

In this work, a robotic platform consisting of different elements was used, including a Romeo BLE control board, an Adafruit HUZZAH32 board and an HC-SR04 ultrasonic sensor. 

The Romeo BLE board from DFRobot is defined as an Arduino-based all-in-one control board specially designed for robotics, which stands out for the possibility of being programmed as if it were an Arduino Uno board and for the integration of Bluetooth 4.0. However, in order to use the latter feature, a special USB adapter, called USB Bluno Link, is required. 

Since this adapter was not available, this advantage could not be exploited in this work, and a board for wireless data communication was added to the system. This board is the HUZZAH32 from Adafruit, an ESP32-based board that supports both Bluetooth technology (both classic and BLE) and WiFi and is also programmable via the Arduino IDE, although thanks to external libraries. 

Finally, the HC-SR04 ultrasonic sensor is used to measure the distance to the nearest object within the measurement range, which will be used to determine whether it is close enough to be considered an obstacle or not. This sensor is a cheap alternative for ultrasonic distance measurement, and theoretically allows distances between 2~cm and 450~cm to be measured with an accuracy in the order of millimeters ($\pm3$ mm). It uses two transducers, one of which emits 8 pulses at a frequency of 40~KHz and the other of which receives the echoes produced by these pulses. This sensor allows to measure the time that has elapsed since these pulses were emitted until they were received in order to make a subsequent conversion to distance using the speed of sound propagation.

The robot is equipped with two different types of power supplies, where one of them powers the motors that allow the movement of the wheels and the Romeo BLE board and the other is destined to power the Adafruit HUZZAH32 board.

\subsection{SpiNNaker}
\label{subsec:spinnaker}

SpiNNaker is a massively-parallel multi-core computing system that was designed to allow modeling very large \acp{SNN} in real time and whose interconnected architecture is inspired by the connectivity characteristics of the mammalian brain \cite{furber2014spinnaker}. 

In this work, a SpiNN-3 machine has been used for the simulation of the \ac{SNN} designed for this work, which can be found in Section~\ref{sec:snn_design}. It has 4 chips, each of them having eighteen ARM968E-S cores operating at 200~MHz. The capacity of the number of neurons that can be computed at the same time during a simulation is slightly limited in this version, but it is more than enough for this work thanks to the low use of resources in the development of the \ac{SNN}. More details about this platform can be found in \cite{rowley2019spinntools}.

\subsubsection{Spiking Neural Networks}
\label{subsubsec:snn}

There are currently considered to be three different generations of \acp{ANN}: classical \acp{ANN}, \acp{DNN} and Spiking Neural Networks (SNNs). Thus, \acp{SNN} are considered the third generation of \acp{ANN}. All \acp{ANN} can be viewed as graphs in which the nodes and edges represent neurons and synapses, respectively. This structure based on artificial models of neurons and synapses, with a high level of abstraction in the case of classical \acp{ANN} and \acp{DNN} \cite{ghosh2009spiking}, is inspired by the biological nervous system. Neuron models used in \acp{SNN} are intended to be as close as possible to the functioning of the biological neurons that can be found in that system, and therefore these \acp{SNN} are considered to be the closest type of \ac{ANN} to their biological counterpart \cite{davidson2021comparison}.

One of the most important aspects of \acp{SNN} is how information is transmitted through the neural network. In the biological nervous system, information is transmitted in the form of asynchronous electrical impulses called spikes, which are large peaks in the membrane potential of biological neurons that occur when the membrane potential exceeds a threshold potential. When these spikes occur in a neuron, they are propagated to all neurons connected to it through synapses, causing or not the generation of new spikes in the target neurons, and so on.

This behavior makes \acp{SNN} more complex than the rest of \acp{ANN}. However, the encoding of information in spikes makes them more energy-efficient as they have to deal with precise timing, which translates into a low computational cost and, therefore, a low power consumption. Some improvements in the hardware implementation of \acp{SNN}, such as avoiding multiplications, processing spikes using shifts and sums, and only transmitting single bits of information instead of real numbers, allow achieving real-time execution \cite{lobo2020spiking}.

\subsubsection{Live injector}
\label{subsubsec:liveinjector}

SpiNNaker supports real-time spike injection, i.e., data injection during simulation using a special type of neuron called spike injector\footnote{\url{https://spinnakermanchester.github.io/2015.004.LittleRascal/InjectingDataRealTime.html}}. It is possible to tell this neuron when to emit a spike in real time. This mechanism is particularly interesting for the development of bio-inspired robotic applications in which it is necessary to convert the information obtained by using digital sensors to spikes so that the neural network can process it.

\subsection{Software}
\label{subsec:software}

The code of this work has been developed using Python for the computer and SpiNNaker, while Arduino has been used for the robotic platform. PyNN \cite{davison2009pynn}, a Python package for the simulator-independent specification of neuronal network models, and sPyNNaker \cite{rhodes2018spynnaker}, an additional Python package which is required to work with PyNN and the SpiNNaker hardware platform, are also used in the computer. Currently, PyNN supports NEURON \cite{hines1997neuron}, NEST \cite{Gewaltig:NEST} and Brian \cite{goodman2008brian} as neural network software simulators, as well as SpiNNaker \cite{furber2014spinnaker} and BrainScaleS neuromorphic hardware systems. PyNN 0.9.6 and sPyNNaker 6.0.0 are used in this work. 

In relation to how the experiments have been carried out to check the correct functioning of the implemented system, Matplotlib 3.6.0 has been used for the representation of graphics.

\section{System description}
\label{sec:system}

\begin{figure}[!ht]
\centering
\includegraphics[width=.5\linewidth]{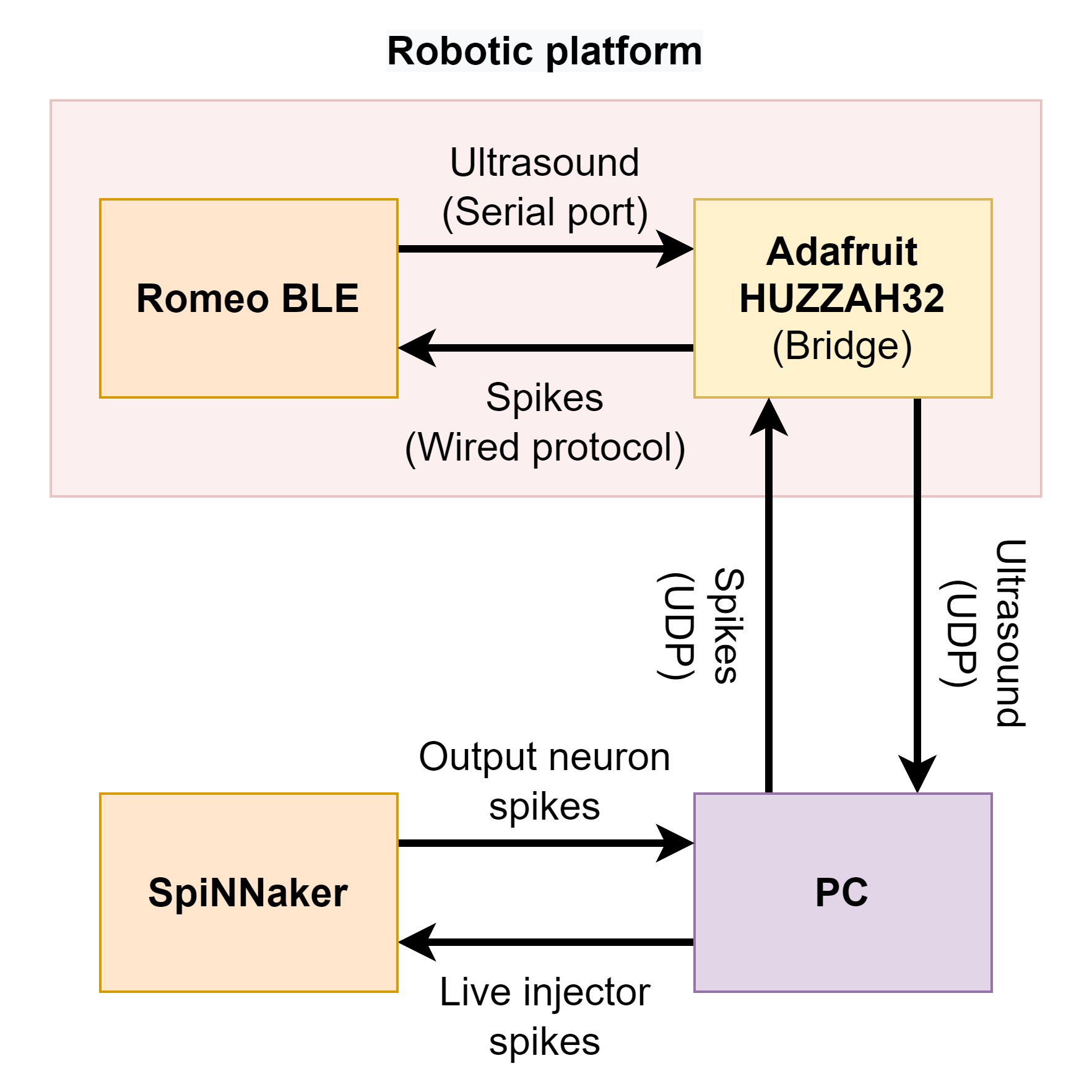}
\caption{General diagram of the complete system. The three main blocks of which it is composed are shown: a robotic platform, the SpiNNaker neuromorphic platform and a computer, together with the type of information that is transmitted between them and the protocols used.}
\label{fig:system_diagram}
\end{figure}

This section describes in depth how the materials presented in Section~\ref{sec:materials_and_methods} are interconnected to form the complete system. Thus, it is composed of three essential blocks, which are as follows: a robotic platform on which the different tests of the implemented obstacle detection system have been carried out, the SpiNNaker neuromorphic platform, which simulated the \ac{SNN} used and whose design is detailed in Section~\ref{sec:snn_design}, and a computer that handled the data exchange between both external platforms and which performs the rest of data computation. These three blocks are shown in the general diagram of the system, presented in Figure~\ref{fig:system_diagram}, which also shows how information is transmitted through it.

\begin{figure}[!ht]
\centering
\includegraphics[width=.5\linewidth]{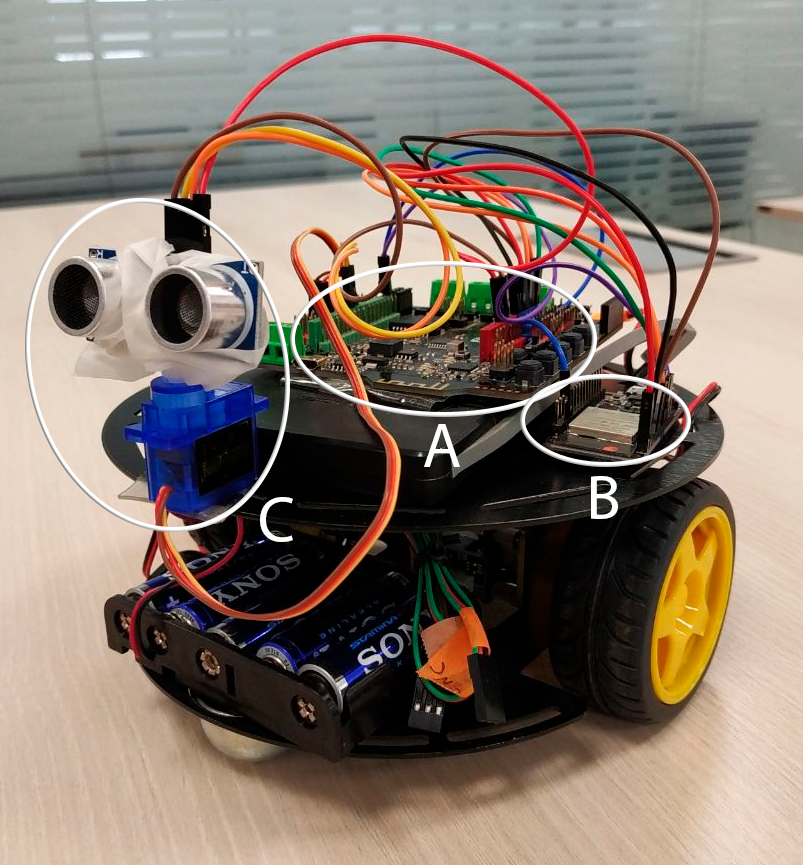}
\caption{Picture of the robotic platform showing its different components: A) Romeo BLE board. B) Adafruit HUZZAH32. C) HC-SR04 ultrasonic sensor.}
\label{fig:robot_photo}
\end{figure}

Figure~\ref{fig:robot_photo} presents the robotic platform used. The most important hardware elements have been highlighted with letters \textit{A}, \textit{B} and \textit{C} for a better identification in the image. \textit{A} and \textit{B} are a Romeo BLE control board and an Adafruit HUZZAH32, respectively. While the Romeo BLE platform takes care of all the computing related to the robotic platform, the latter is just a bridge for proper communication over UDP with the computer. Both have been programmed using Arduino. Finally, \textit{C} corresponds to an ultrasonic sensor mounted on a servomotor that allows rotary movements. However, the servomotor was not used in this work and the ultrasonic sensor always pointed in a fixed direction (forward).

Obstacle detection starts at this point. Initially, the robot is continuously reading the distance values thanks to the ultrasonic sensor. These values are transmitted from the Romeo BLE board to the Adafruit HUZZAH32 via the serial port, and then to the computer via UDP. 

At this point, the computer is responsible for generating a spike train that changes according to the received values, which are the measurements obtained from the ultrasonic sensor. This spike train is sent to and processed by SpiNNaker, and then the result is returned via the computer to the Adafruit HUZZAH32 board in the form of UDP packets representing the spikes of the output neuron of the implemented \ac{SNN}, whose design is also shown in Section~\ref{sec:snn_design}. The spikes are then transmitted to the Romeo BLE via the Adafruit HUZZAH32 thanks to a simple wired protocol.

When these spikes arrive back to the robotic platform, the Romeo BLE is expected to execute the commands of a collision avoidance algorithm. Since collision avoidance is outside the scope of this work, it has been implemented by simply turning to the right until no obstacles are detected, which implies no spike reception for a period of 500~ms. This value could be reduced, although it is convenient to use relatively high times to allow the transmission of information, especially because of the delays produced by the use of WiFi.

Another important aspect to detail in this section is how the measurements are obtained thanks to the ultrasonic sensor. Since the system is intended to be reliable in order to avoid false positives when detecting obstacles, and also knowing that there are a large number of cases in which erroneous measurements can be obtained due to the way the sensing is performed using ultrasounds, an algorithm has been implemented with the aim of increasing this reliability by means of measurement redundancy. This, on the other hand, increases the computational and temporal cost of the system, but not enough to seriously affect its real-time capability. This algorithm, whose pseudocode is shown in Algorithm~\ref{alg:ultrasounds}, is based on setting a number of measurements, $maxHits$, to be produced redundantly, which means that these measurements are within a margin of error from the first measurement taken as the reference measurement. If all the measurements are within this margin, this first measurement is taken as correct and it is transmitted to the computer. Otherwise, i.e., when a measurement is found that is not within this margin, it is taken as the new reference measurement and the process is repeated for the next $maxHits$ measurements.

In this work, we considered $maxHits$ to be equal to 4 and a maximum error of 120~ms (approximately 2~cm) in the calculation of ultrasonic measurements. If $maxHits$ were lower than 4, the speed with which the effective measurements are obtained would be increased, which would result in a small increase in the real-time capability of the system, but the reliability of the resulting measurement would be decreased. Increasing $maxHits$ would have the opposite effect. On the other hand, the value set as maximum error should be more than enough to obtain correct measurements, since the HC-SR04 ultrasonic sensor has a theoretical accuracy in the order of millimeters ($\pm3$ mm).

\begin{algorithm}

\caption{Algorithm for increasing the reliability of ultrasonic measurements. $a$ and $b$ represent natural numbers.}\label{alg:ultrasounds}

\begin{algorithmic}
\State $hits \gets 0$
\State $maxHits \gets a$ 
\State $maxError \gets b$
\While{$True$}
    \If{$hits == 0$} \Comment{Taking of the first measurement}
        \State $firstMeasurement \gets$ \textbf{Obtain a new measurement} $\dots$ 
        \State $hits \gets 1$
    \ElsIf{$hits < maxHits$} \Comment{Taking of the following measurements}
        \State $newMeasurement \gets$ \textbf{Obtain a new measurement} $\dots$ 
        \If{$|newMeasurement - firstMeasurement| \leq maxError$}
            \State $hits \gets hits + 1$
        \Else \Comment{Restart without sending}
            \State $firstMeasurement \gets newMeasurement$ 
            \State $hits \gets 1$
        \EndIf
    \Else \Comment{Send and restart}
        \State \textbf{Send firstMeasurement to the computer} $\dots$
        \State $hits \gets 0$
    \EndIf
\EndWhile
\end{algorithmic}
\end{algorithm}
\section{Spiking Neural Network}
\label{sec:snn_design}

\begin{figure}[!ht]
\centering
\includegraphics[width=.5\linewidth]{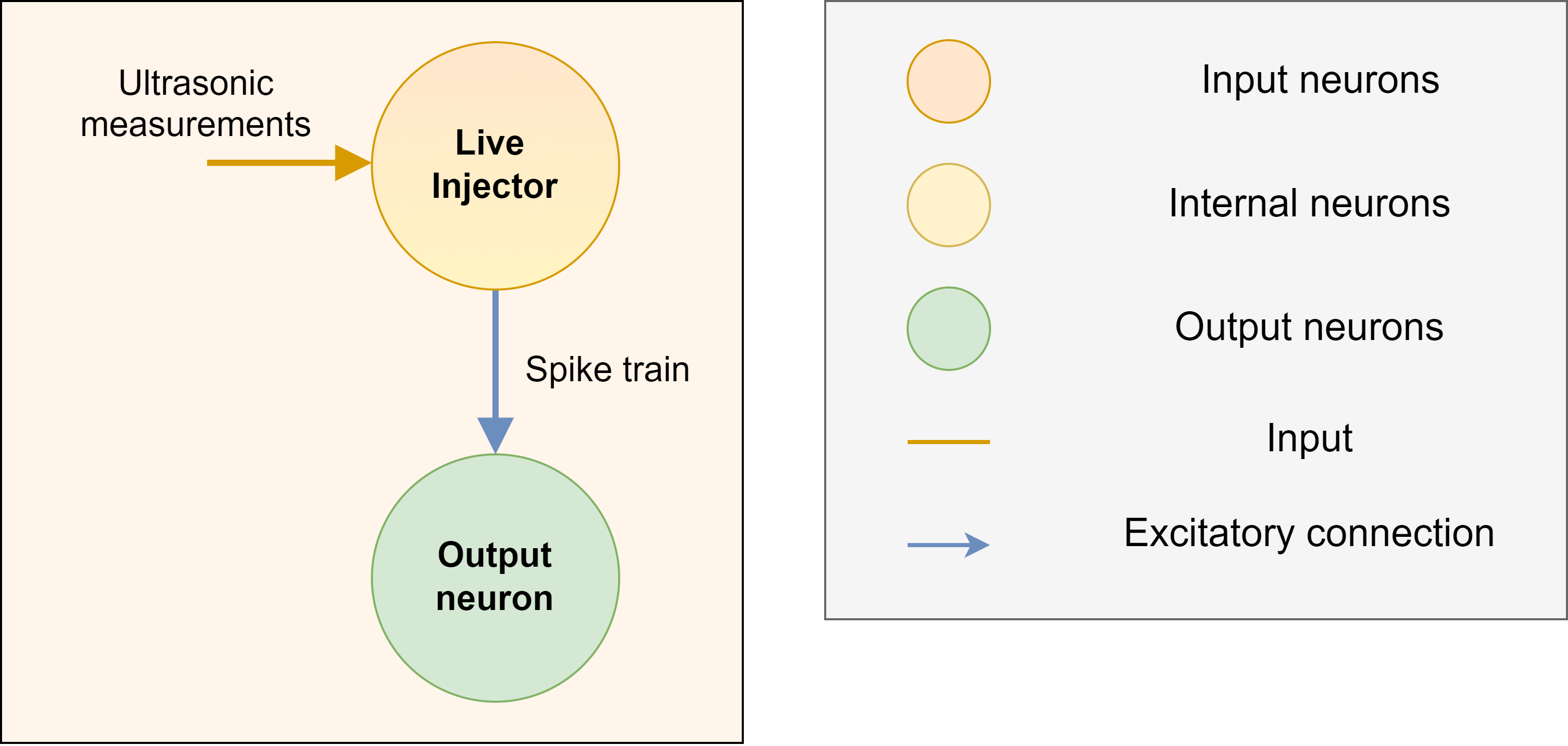}
\caption{General diagram of the implemented \ac{SNN}.}
\label{fig:snn_diagram}
\end{figure}

This section provides information on the structure of the \ac{SNN} used, whose design is shown in Figure~\ref{fig:snn_diagram}. This neural network is implemented by using two neurons: the live injector mentioned in Section~\ref{sec:materials_and_methods} and an output neuron, resulting in a low-latency model that requires the use of very few resources. The behavior of the output neuron is defined by the \ac{LIF} neuron model. An excitatory synapse with a delay of 1 time step (1~ms in this case, since it is the default time step in sPyNNaker) is used to connect the live injector to the output neuron.

Both neurons in the \ac{SNN} have very specific functions which must be studied in depth to understand how obstacle detection is performed in the system. Thus, a different subsection is dedicated to each of the neurons. While the live injector is in charge of generating a frequency-variable spike train from the digital information obtained from the ultrasonic sensor, the output neuron has the function of processing this spike train to be able to decide whether an object is too close (it is considered an obstacle) or not.

\subsection{On how to encode digital information into spike trains}
\label{subsec:spike_train_encoding}

In a neuron, there are two basic states: one in which the neuron does not fire, i.e., the membrane potential is below the threshold potential, and one in which it does. Both states could be associated to Boolean values by the absence (0 or false) or existence (1 or true) of an output spike fired at a given time instant. Note that neurons should be in the first state most of the time, following the principle of low power consumption of \acp{SNN}. 

Both states are directly related to spikes, and then, they are important to understand how information is encoded when using \acp{SNN}.
In \cite{auge2021survey}, several existing encoding methods in \acp{SNN} are deeply detailed from a biological point of view. These methods are useful for converting digital information into spiking information, each of them having a series of advantages and disadvantages that make it more suitable for specific cases. For example, an array of neurons could be used to fire according to the binary encoding of a digital number, which would imply using one neuron for each bit in the binary encoding. In addition, it would be necessary to synchronize their output spikes to ensure that they are part of the same representation of the input number, and not that of an earlier or later encoded number. Thus, this method would be related to temporal coding. However, the number of neurons used could be reduced by using a single neuron whose firing rate encodes that number, which would be related to rate coding.

Rate coding is used in this work to reduce the amount of resources (neurons and synapses) required, providing a low-cost solution for obstacle detection task using \acp{SNN}. The live injector shown in Figure~\ref{fig:snn_diagram} is responsible for encoding the information obtained by the ultrasonic sensor into spikes. In this way, every time a new data (the \ac{ToF}, i.e., the time elapsed since an ultrasonic wave is sent from the ultrasonic sensor until it is received after bouncing off the obstacle) arrives to the computer, it is used to calculate the new \ac{ISI} of the live injector, that refers to the time that must pass between two generated spikes. Simultaneously, the computer also calculates the actual time difference between when the last spike was generated (or fired) and the current time. When this value is greater than or equal to the calculated \ac{ISI}, the live injector must fire again. The firing rate is the inverse of this actual time difference, and it should be approximately equal to the inverse of the calculated \ac{ISI}. The \ac{ISI}, in seconds, is calculated using the formula presented in Equation~\ref{formula:li_ISI}.

\begin{equation}
    \label{formula:li_ISI}
    \ac{ISI} = \left( \frac{x}{5883}\right)^{\!2} + 0.001
\end{equation}

In this formula, $x$ refers to the \ac{ToF}, in microseconds, provided by the ultrasonic sensor and is manually limited to 5883~$\mu$s as maximum, which is approximately the \ac{ToF} of an ultrasonic wave that bounces at a distance of 100~cm from the ultrasonic sensor. Distances greater than 100~cm are limited to this value to reduce the range of distances to be considered in the formula. Note that a distance of 100~cm would not make the system to fire. In this way, the first term of the equation allows the values of the calculated \ac{ISI} to be delimited in the range [0, 1] seconds. Having a delimited range of values is critical in the development of real-time robotic systems since they must be inherently deterministic. This term is squared to provide greater differentiation of the calculated \acp{ISI}, especially for high and intermediate $x$ values. In order to avoid unexpected behaviors of the system, a second term has been added to ensure that the minimum \ac{ISI} is 1~ms, which is the simulation time step. Therefore, the calculations are bounded between 0.001 seconds and 1.001 seconds, which are associated to firing frequencies that range between 1000~Hz and 1~Hz, respectively.

\subsection{On how spike trains are processed by the system}
\label{subsec:spike_train_processing}

Each of the spikes generated by the live injector, following the process explained in the previous subsection, is part of a frequency-variable spike train that propagates to the output neuron through an excitatory synapse, as shown in Figure~\ref{fig:snn_diagram}. As already discussed, this output neuron has the function of detecting the presence of an obstacle based on that spike train. The complexity of this task is related to the adjustment of the distance from which an object can be considered as an obstacle, which will be called threshold distance from now on. To adjust this threshold distance and, in general, to process the input spike train, it is necessary to modify the parameters used for the output neuron. The idea was to make the output neuron more or less sensitive to input stimuli, so that all distances above that threshold distance have an associated firing rate in the spike train that is not sufficient to cause the output neuron to fire. Due to the nature of the application developed, in which a mobile robot must perform obstacle detection to avoid collisions, the initial goal in adjusting the neuron parameters was to achieve a threshold distance between 30~cm and 50~cm.

\begin{table}
    \centering
    \caption{Set of neuron parameters used for the output neuron.}
    \label{neuron_parameters}
    \begin{tabular}{|c|c|c|}
    \hline
    
    \textbf{Neuron parameter} & \textbf{Overview} & \textbf{Value} \\ 
    
    \hline
    $c_m$ & Membrane capacitance & 1.0 nF \\
    
    \hline
    $tau_m$ & Time-constant of the RC circuit & 100.0 ms \\
    
    \hline
    $tau_{refrac}$ & Refractory period & 0.0 ms \\
    
    \hline
    $tau_{syn\_E}$ & Excitatory input current decay time-constant & 5.0 ms \\
    
    \hline
    $tau_{syn\_I}$ & Inhibitory input current decay time-constant & 5.0 ms \\
    
    \hline
    $v_{rest}$ & Resting potential & -65.0 mV \\
    
    \hline
    $v_{reset}$ & Reset potential & -65.0 mV \\
    
    \hline
    $v_{thresh}$ & Threshold potential & -59.5 mV \\
    
    \hline
    \end{tabular}
\end{table}

Table~\ref{neuron_parameters} contains the parameters used for the output neuron. These parameters are almost the same of a default \ac{LIF} neuron with fixed threshold and decaying-exponential post-synaptic current of PyNN\footnote{\url{https://neuralensemble.org/docs/PyNN/reference/neuronmodels.html\#pyNN.standardmodels.cells.IF_curr_exp}} but with three differences. These differences are relevant for several reasons, which can be explained through the equations of the \ac{LIF} neuron model presented in sPyNNaker \cite{rhodes2018spynnaker} and are as follows:

\begin{enumerate}
    \item $tau_m$ has been increased from 20~ms to 100~ms. This decreases the absolute value of $dV/dt$, increasing the time it takes for the membrane potential to reach its resting potential again from an excited state, i.e., the duration of the repolarization and hyperpolarization phases. Given that each input stimulus produces an increase in membrane potential (depolarization), increasing this duration implies that a lower input firing rate is required to cause an overlap between the repolarization phase of one input stimulus and the depolarization phase produced by the next. Such an overlap would cause the depolarization phase to begin at a point where the membrane potential is above the resting potential. This overlap is key to understanding the behavior of the neural network and to understanding how obstacle detection is performed.

    \item $tau_{refrac}$ has been set to 0~ms to allow the output neuron to fire each time step of the simulation. In this way, the neuron is able to react instantly to the input stimuli.

    \item $v_{thresh}$ has been decreased from -50.0~mV to -59.5~mV. In line with what was explained above for $tau_m$, the decrease of the threshold potential could be considered as an adjustment to make it easier to cause the neuron to fire after the overlap of the repolarization and depolarization phases. While this overlap depends on $tau_m$ and the frequency of the input stimuli, causing the neuron to fire by means of decreasing $v_{thresh}$ also depends on the membrane potential increase produced by each of the input stimuli. This change is intended to increase the number of output spikes fired by the output neuron, which is of great interest to improve the real-time capability of the system.
\end{enumerate}

To understand how this spike train processing works, an in-depth theoretical study of the relationship between the frequency of the input stimuli of a neuron and the output spikes fired by that neuron, as well as its membrane potential, has been carried out. This study focuses on the effect of the overlap between the repolarization and depolarization phases.

\begin{figure}[!ht]
\centering
\includegraphics[width=\linewidth]{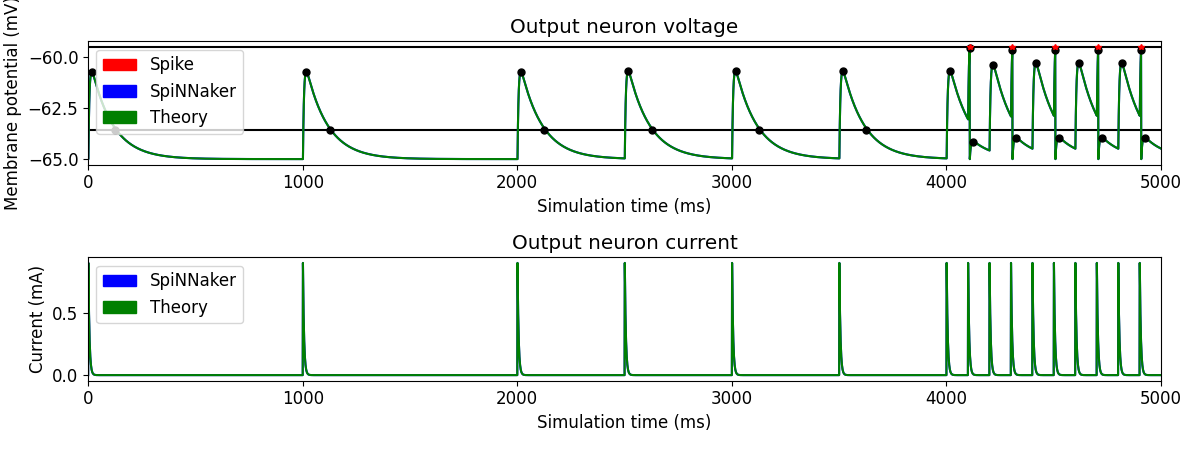}
\includegraphics[width=\linewidth]{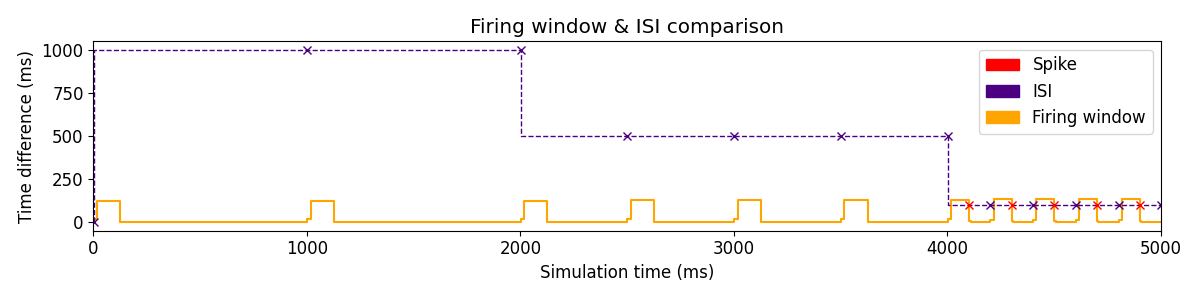}
\caption{Example of output neuron response to an input spike train. At the top of the figure, the membrane potential calculated using the equations presented in \cite{rhodes2018spynnaker} and the membrane potential obtained from the SpiNNaker hardware platforms after simulation are compared. The upper black line delimits the threshold potential, while the lower black line delimits the minimum potential necessary to fire an output spike when an input spike is received. At the middle, the synaptic currents are shown. At the bottom, a comparison of the ISI of input spikes and the firing windows is made. Input spikes are marked with magenta crosses if they do not cause the output neuron to fire, or with red crosses if they do.}
\label{fig:spinnaker_comparison_fig1}
\end{figure}

Figure~\ref{fig:spinnaker_comparison_fig1} shows an example in which the output neuron is provided with an input spike train containing three different firing rates. These are the following: 1) 1~Hz from 0 ms to 2000~ms. 2) 2~Hz from 2000~ms to 4000~ms. 3) 10~Hz from 4000~ms to 5000~ms. 

The upper graph shows a comparison between the theoretical membrane potential of the output neuron using the equations presented in \cite{rhodes2018spynnaker} and the empirical one obtained from the SpiNNaker hardware platforms after simulation. Note that both the theoretical and empirical membrane potentials are exactly the same since these equations are also used by SpiNNaker. The upper black line delimits the threshold potential, while the lower black line delimits the minimum potential that the neuron must have so that, when receiving an input spike, the neuron fires. The middle graph shows the synaptic currents induced in the output neuron in response of the arrival of input spikes.

It should be noted that, having an excitatory synapse with weight equal to 1 nA connecting the live injector to the output neuron, it is theorized that the synaptic current should be increased by 1 mA each time an input spike arrives. However, as explained in \cite{rhodes2018spynnaker}, SpiNNaker uses the Euler's method to solve the differential equations of the \ac{LIF} model and, in order to correct the intrinsic cumulative error of this solution, synaptic currents are decayed. Thus, there is a small difference from the theoretical value of these currents and they have to be measured after simulation. In this way, the induced synaptic current by each input spike is approximately 0.9063~mA. Using this current value, the parameters used for the output neuron shown in Table~\ref{neuron_parameters} and the theoretical equations, it is possible to calculate the value of the minimum firing potential, which is approximately equal to -63.569~mV.

Receiving an input spike in a state in which the membrane potential of the neuron is above this minimum firing potential causes the neuron to fire, producing an output spike. This is why this minimum firing potential is key to understanding how the input spike train is processed. 

In order to explain how the output neuron behaves, it is of great interest to calculate the time during which the neuron is able to fire. From now on, this time frame will be called firing window. There are two simple cases for which the calculation of this firing window is done in different ways:

\begin{enumerate}
    \item When an input spike arrives, the induced synaptic current ensures that, at some point, a maximum membrane potential ($dV/dt = 0$, $V \neq V_{rest}$) is reached that is above the minimum firing potential. Although the membrane potential may be below the minimum firing potential at the time the input spike arrives, the induced synaptic current is sufficient to cause the neuron to fire upon the arrival of another input spike even before the minimum firing potential is reached. Therefore, in this case, the firing window is calculated as the time difference between the arrival of the input spike and the reaching of the associated maximum potential.
    
    \item Another firing window is calculated from the time difference between the reaching of the maximum potential and, after that, the reaching of the minimum firing potential.
\end{enumerate}

The spike train used in Figure~\ref{fig:spinnaker_comparison_fig1} does not consider complex cases in order to facilitate the reader's understanding, since these are not within the scope of this work. In the lower graph, firing windows are shown in orange. The \ac{ISI} of the input spikes is also shown in magenta. While red crosses indicate that an output spike was fired after receiving an input spike, magenta crosses indicate that no output spike was fired. In this way, it should be noted that output spikes were fired only when the \ac{ISI} of the input spikes was lower than the firing window. When output spikes were fired, the firing window was modified.

Thus, this example shows how the output neuron is not able to fire when an input spike train is provided with a firing rate of 1~Hz or 2~Hz, but it does when the firing rate is 10~Hz, since the value of the \acp{ISI} becomes lower than the current firing window. As explained in Section~\ref{subsec:spike_train_encoding}, higher firing rates are associated with objects more closely located to the ultrasonic sensor, which are more likely to be considered obstacles.

\section{Experimentation and results}
\label{sec:experiments_and_results}

Different experiments were carried out to test the performance of the system. These experiments can be classified into different types, depending on what aspect of the system was intended to be tested. In this section, some of the most relevant are explained in depth and their results are presented to verify and highlight the validity of the system for real-time obstacle detection. The list of experiments presented in this section is as follows:

\begin{enumerate}
    \item Two tests to confirm the value of the threshold distance.

    \item One test to verify the response of the output neuron to increasing and decreasing distances.

    \item Two tests to prove the real-time capability of the system.

    \item One test to prove the correct functioning of the complete system in real environments.
\end{enumerate}

All these experiments start with a default \ac{ToF} of 5883 $\mu s$, which corresponds to a distance of approximately 100 cm between the sensor and the object, meaning that when this value appears in the graphs the computer had not yet received any measurements.

Note that the results of these experiments show unexpected temporary gaps between the spikes of the output neuron. The current state of the membrane potential plays a key role here, in line with what is explained in Section~\ref{subsec:spike_train_processing}. In this way, although there is a straightforward relationship between the firing rate of the input spike train and the firing rate of the output neuron, there are small differences in the rates (i.e., these gaps) which are produced because of how the membrane potential of the output neuron is always varying in response to input spikes.

\subsection{Threshold distance}

As previously explained in this paper, the threshold distance depends on the parameters used for the output neuron. This experiment aimed to verify that the threshold distance is within the desired range (between 30~cm and 50~cm, in this case) by using artificial measurements generated by the Romeo BLE board that simulate the \ac{ToF} values obtained thanks to the ultrasonic sensor, and which are related to the distance between the sensor and the object. Multiple tests were performed for multiple distances between 10~cm and 50~cm. 

In particular, the threshold distance could be approximated by finding that, for a certain distance, output spikes were still fired, but for the next further distance this was no longer the case. Specifically, these distances were 39 cm and 39.5 cm, respectively. This guarantees that the threshold distance is between these two values, so it could be approximated to 39 cm.

\begin{figure}[!ht]
\centering
\includegraphics[width=\linewidth]{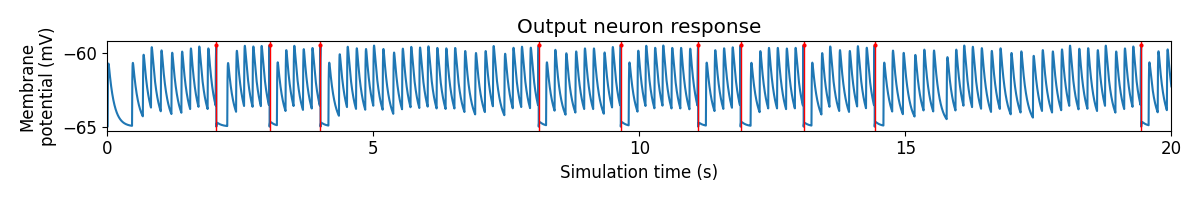}
\includegraphics[width=\linewidth]{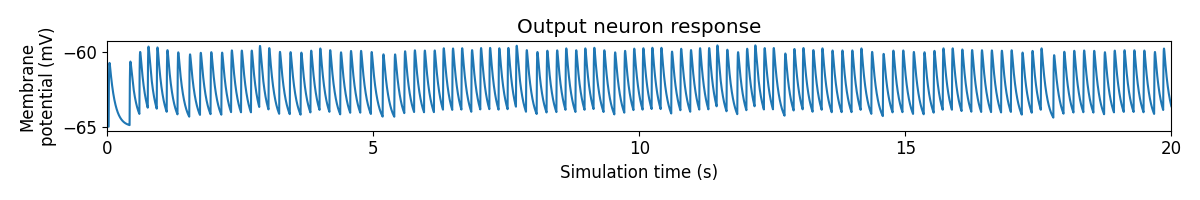}
\caption{System response to measurements sent by the Romeo BLE board corresponding to an object located at a constant distance of approximately 39 cm (top) and 39.5 cm (bottom). Output spikes are marked with red points and vertical lines.}
\label{fig:threshold_test}
\end{figure}

Figure~\ref{fig:threshold_test} shows the output neuron response for both cases. The graph at the top shows that, for a distance of 39~cm, the neuron was still firing spikes. However, the \ac{ISI} of these output spikes was generally high because it took longer to reach sufficient excitation to cause each spike. Increasing the distance to the object decreases the frequency of the input spike train, being the overlap between repolarization and depolarization phases a bit smaller, so there is a point where it is not enough to cause the output neuron to fire. This is what happens just between the distances of 39 and 39.5~cm, as can be seen for the latter case shown in the graph at the bottom of Fig.~\ref{fig:threshold_test}.

\subsection{Increasing and decreasing distances}

In this experiment, different measurements were also artificially sent from the Romeo BLE board. However, these \ac{ToF} values increased and decreased over time, which was directly related to increasing and decreasing the distance between the sensor and the obstacle. Initially, there is a part where the values increased and decreased linearly. Then the measurements changed abruptly, with larger or smaller jumps in the values. This last part will be the focus of the next experiment.

\begin{figure}[!ht]
\centering
\includegraphics[width=\linewidth]{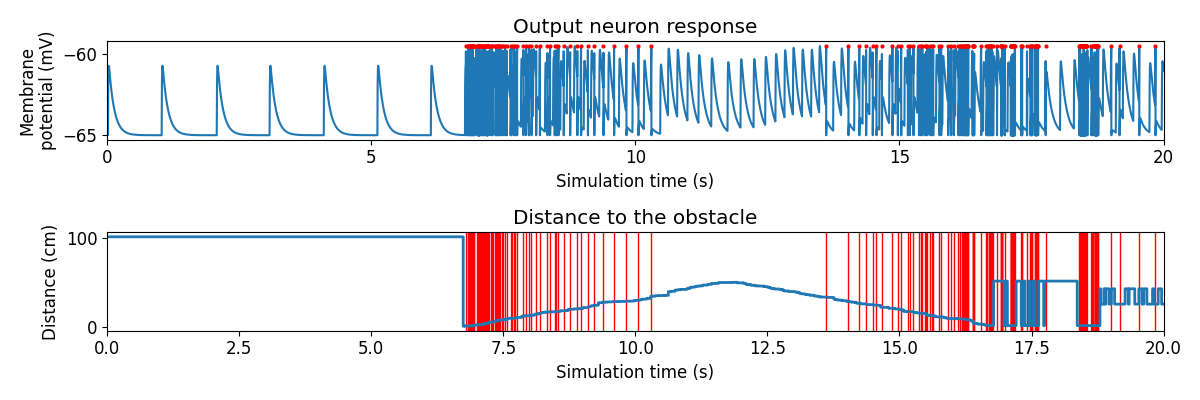}
\caption{System response to time-varying distances at which the object is placed from the ultrasonic sensor. Output spikes are marked with red points and vertical lines.}
\label{fig:increasing_decreasing_test}
\end{figure}

Figure~\ref{fig:increasing_decreasing_test} shows the results of this experiment. As it can be observed, for low \ac{ToF} values (from 60~$\mu s$) the firing rate of the output neuron was high. As this distance increased (up to 3000~$\mu s$), the firing rate decreased. Another of the most interesting aspects of this experiment is that it can be clearly observed how upon reaching a minimum firing rate, associated to the threshold distance, this firing rate became insufficient to make the neuron firing, as explained in the first experiment. This is why a region appears in which no spikes were fired and the membrane potential of the neuron started to decrease, as distance was being increased above the threshold distance. After that, the reverse process occurs, with the distance decreasing and the firing rate increasing sufficiently for the repolarization/depolarization overlap to cause the neuron to start firing again.

\subsection{Real-time capability}

\begin{figure}[!ht]
\centering
a)
\includegraphics[width=\linewidth]{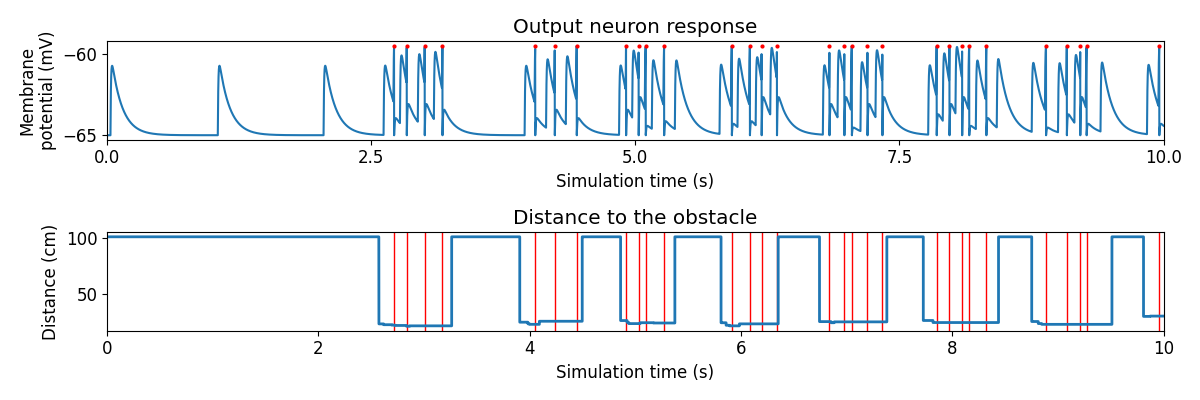}
b)
\includegraphics[width=\linewidth]{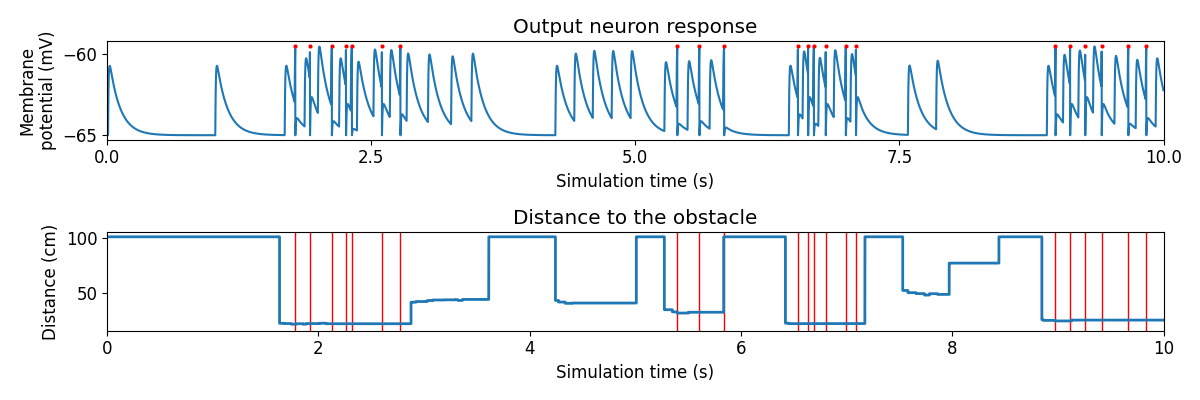}
\caption{System response to appearing and disappearing objects placed at a constant distance (a) and placed at different distances (b). Output spikes are marked with red points and vertical lines.}
\label{fig:real_time_test}
\end{figure}

The purpose of this experiment is to ensure that the system is able to react quickly enough to objects that appear spontaneously in front of the robot. A study of the real-time capability of the system involves not only checking that the reaction time is low enough not to affect the behavior of the system, but also checking that the reaction time is deterministic, i.e., that it is within a range or bounded.

Figure~\ref{fig:real_time_test} shows the result for two tests of this experiment. The two graphs at the top shows the results for a test in which the object appeared at a certain distance from the ultrasonic sensor, while the two graphs at the bottom shows the results for a test in which different distances were being tested.

In the first case, the distance at which the object appeared was below the threshold distance, so whenever the object appeared the system fired output spikes, meaning that the object had been considered an obstacle. Because the appearing distance (about 25~cm) does not correspond to the closest distance at which the object could be found, the calculated \acp{ISI} for the generation of the input spike train, which were around 60~ms, were not the lowest. Thus, the firing rate is not particularly high. In this way, the \ac{ISI} of the output spikes range from approximately 60 ms (which would be the minimum, due to the fact that it is the \ac{ISI} of the input spike train) to approximately 130 ms.

In the case of the graph at the bottom, it can be seen in more detail how for longer distances lower firing rates were achieved. This is directly related to a lower repolarization/depolarization overlap in membrane potential produced in response to input spikes. In addition, there were different distance levels at which no output spikes were fired, meaning that these distances were above the threshold distance.

\subsection{Real environment}

This experiment was intended to verify that the complete system worked as expected in a real environment. The robotic platform was positioned in the center of an area surrounded by rectangular cardboard boxes. 

The complete system is defined so that the robotic platform moves forward as long as no obstacles are detected and, if obstacles are detected, it turns to the right until obstacles are no longer detected.

\begin{figure}[!ht]
\centering
\includegraphics[width=\linewidth]{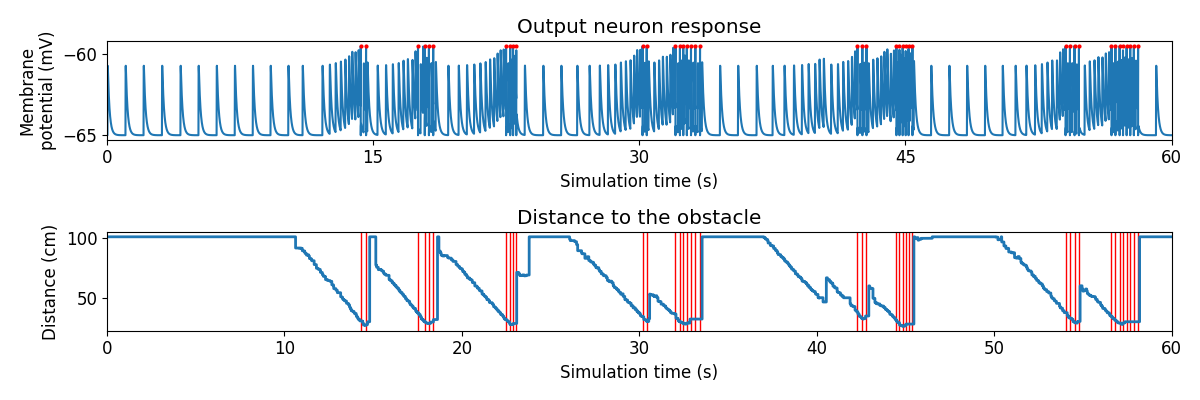}
\caption{System response in a real environment. Output spikes are marked with red crosses and vertical lines.}
\label{fig:real_test}
\end{figure}

The results obtained during the execution of one of the tests of this experiment are shown in Figure~\ref{fig:real_test}. In this test, it can be seen how the system started by measuring an obstacle at a distance of 100~cm or more. Over time, as the robotic platform moved forward, this distance was reduced, which can be observed as a linear decrease in distance. When this distance dropped below the threshold distance, the system detected the obstacle (output spikes appeared) and sent the command to the robotic platform to start turning to the right until the obstacle was no longer detected. At that point, the output neuron stopped firing spikes and the distance measured by the ultrasonic sensor increased, with this new measurement corresponding to the distance between the ultrasonic sensor and the closest object it could find in the new direction it was pointing.


\section{Discussion and future work}
\label{sec:discussion}

The results obtained and shown in Section~\ref{sec:experiments_and_results} seem to be good enough to validate the implemented system. The study of the firing windows seems to be useful to understand to some extent how and why the output neuron responds to certain input stimuli and how it relates to the \ac{ISI}. In this way, the implemented \ac{SNN}, whose essence lies in the functioning of the LIF neuron, works as a high-pass spiking filter, where the \ac{SNN} does not respond to low frequencies of the input spike train. This is really interesting since it could be the basis for the study of new applications based on spiking filters using this approach.

To improve the results obtained in this work, it could also be interesting to study the implementation of the proposed \ac{SNN} in other neuromorphic platforms, with the purpose of minimizing the ISI of the input spike train (adjusting at the same time the presented formula in Section~\ref{subsec:spike_train_encoding} to these values), which would allow to obtain a better performance of the system for the obstacle detection task. 

Since the robotic platform is an independent block within the complete system, any modification can be made to it. This means that it would be possible to increase the number of range sensors of the same type used for the obstacle detection task, as long as an algorithm is implemented so that the measurement sent to the central computer is unique, or the network architecture is replicated for each of the sensors. On the other hand, it is also possible to use different types of range sensors. This is interesting because one of the possible modifications to the system could be to use infrared sensors to support the ultrasonic sensors, whereby it would be possible to try to smooth out or eliminate the problems inherent in ultrasonic sensors, especially with regard to the sound reflection angles that prevent correct measurements in certain cases. Moreover, this second sensor should help to increase the reliability of the obtained measurements.

Section~\ref{subsec:spike_train_processing} discusses some fundamental details to understand why the parameters used for the output neuron have been chosen. Thus, it is explained that the decrease in the threshold potential of the neuron is important to increase the number of spikes produced in the output response of the system, which is directly related to its performance in real time. It is true that this would imply an increase in the power consumption of the network, despite the fact that it is generally tried to squeeze as much as possible out of the low power consumption of \acp{SNN}.
However, sometimes the priority is to increase the performance of the entire system, which requires an increase in power consumption, so a decision would have to be made in the future (when refining such an application) on how much to increase performance without increasing power consumption too much, finding a balance. This debate has been present throughout the history of computers.

\section{Conclusions}
\label{sec:conclusions}

This paper discusses the interest of obstacle detection from the perspective of robotics and neuroscience, and propose the implementation of a \ac{SNN}-based system to perform obstacle detection to exploit the advantages of these bio-inspired architectures. An in-depth explanation of the functioning of the implemented \ac{SNN} is given, containing details regarding how the input spike train is generated and how it is processed to perform obstacle detection in the output neuron. It also contains an in-depth analysis of how the \ac{ISI} of this input spike train is related to the output firing rate of the system.

A series of experiments were carried out from different approaches: the threshold distance bounding, the response to increasing and decreasing distances, the response to sudden distance jumps and, finally, the functioning of the system in a real environment. The results obtained for each of the experiments validates the implemented system. In this way, it is stated how the output firing rate is related to the distance measured by the ultrasonic sensor, increasing when it decreases and vice versa. 

In order to understand in detail the functioning of the implemented system, an in-depth study of how information is encoded and processed is carried out. It has been discussed how this analysis could be very useful for the development of applications based on spiking filters.

\medskip
\textbf{Acknowledgements} \par 
This research was partially supported by the Spanish project MINDROB (PID2019-105556GB-C33) funded by MCIN/AEI/10.13039/501100011033. D. C.-M. was supported by a ``Formaci\'{o}n de Profesor Universitario'' Scholarship from the Spanish Ministry of Education, Culture and Sport.

\bmhead{Conflict of interest}

The authors have no conflicts of interest to declare that are relevant to the content of this article.

\bmhead{Data availability}

Not applicable.

\bmhead{Ethics approval}

Not applicable.

\bmhead{Consent to participate}

Not applicable.

\bmhead{Consent for publication}

Not applicable.

\bibliography{sections/bibliography}


\begin{thebibliography}{34}
\ifx \bisbn   \undefined \def \bisbn  #1{ISBN #1}\fi
\ifx \binits  \undefined \def \binits#1{#1}\fi
\ifx \bauthor  \undefined \def \bauthor#1{#1}\fi
\ifx \batitle  \undefined \def \batitle#1{#1}\fi
\ifx \bjtitle  \undefined \def \bjtitle#1{#1}\fi
\ifx \bvolume  \undefined \def \bvolume#1{\textbf{#1}}\fi
\ifx \byear  \undefined \def \byear#1{#1}\fi
\ifx \bissue  \undefined \def \bissue#1{#1}\fi
\ifx \bfpage  \undefined \def \bfpage#1{#1}\fi
\ifx \blpage  \undefined \def \blpage #1{#1}\fi
\ifx \burl  \undefined \def \burl#1{\textsf{#1}}\fi
\ifx \doiurl  \undefined \def \doiurl#1{\url{https://doi.org/#1}}\fi
\ifx \betal  \undefined \def \betal{\textit{et al.}}\fi
\ifx \binstitute  \undefined \def \binstitute#1{#1}\fi
\ifx \binstitutionaled  \undefined \def \binstitutionaled#1{#1}\fi
\ifx \bctitle  \undefined \def \bctitle#1{#1}\fi
\ifx \beditor  \undefined \def \beditor#1{#1}\fi
\ifx \bpublisher  \undefined \def \bpublisher#1{#1}\fi
\ifx \bbtitle  \undefined \def \bbtitle#1{#1}\fi
\ifx \bedition  \undefined \def \bedition#1{#1}\fi
\ifx \bseriesno  \undefined \def \bseriesno#1{#1}\fi
\ifx \blocation  \undefined \def \blocation#1{#1}\fi
\ifx \bsertitle  \undefined \def \bsertitle#1{#1}\fi
\ifx \bsnm \undefined \def \bsnm#1{#1}\fi
\ifx \bsuffix \undefined \def \bsuffix#1{#1}\fi
\ifx \bparticle \undefined \def \bparticle#1{#1}\fi
\ifx \barticle \undefined \def \barticle#1{#1}\fi
\bibcommenthead
\ifx \bconfdate \undefined \def \bconfdate #1{#1}\fi
\ifx \botherref \undefined \def \botherref #1{#1}\fi
\ifx \url \undefined \def \url#1{\textsf{#1}}\fi
\ifx \bchapter \undefined \def \bchapter#1{#1}\fi
\ifx \bbook \undefined \def \bbook#1{#1}\fi
\ifx \bcomment \undefined \def \bcomment#1{#1}\fi
\ifx \oauthor \undefined \def \oauthor#1{#1}\fi
\ifx \citeauthoryear \undefined \def \citeauthoryear#1{#1}\fi
\ifx \endbibitem  \undefined \def \endbibitem {}\fi
\ifx \bconflocation  \undefined \def \bconflocation#1{#1}\fi
\ifx \arxivurl  \undefined \def \arxivurl#1{\textsf{#1}}\fi
\csname PreBibitemsHook\endcsname

\bibitem{tzafestas2013introduction}
\begin{bbook}
\bauthor{\bsnm{Tzafestas}, \binits{S.G.}}:
\bbtitle{Introduction to Mobile Robot Control}.
\bpublisher{Elsevier}, \blocation{???}
(\byear{2013})
\end{bbook}
\endbibitem

\bibitem{hu2020survey}
\begin{barticle}
\bauthor{\bsnm{Hu}, \binits{J.-w.}},
\bauthor{\bsnm{Zheng}, \binits{B.-y.}},
\bauthor{\bsnm{Wang}, \binits{C.}},
\bauthor{\bsnm{Zhao}, \binits{C.-h.}},
\bauthor{\bsnm{Hou}, \binits{X.-l.}},
\bauthor{\bsnm{Pan}, \binits{Q.}},
\bauthor{\bsnm{Xu}, \binits{Z.}}:
\batitle{A survey on multi-sensor fusion based obstacle detection for intelligent ground vehicles in off-road environments}.
\bjtitle{Frontiers of Information Technology \& Electronic Engineering}
\bvolume{21}(\bissue{5}),
\bfpage{675}--\blpage{692}
(\byear{2020})
\end{barticle}
\endbibitem

\bibitem{john2019rvnet}
\begin{bchapter}
\bauthor{\bsnm{John}, \binits{V.}},
\bauthor{\bsnm{Mita}, \binits{S.}}:
\bctitle{Rvnet: Deep sensor fusion of monocular camera and radar for image-based obstacle detection in challenging environments}.
In: \bbtitle{Pacific-Rim Symposium on Image and Video Technology},
pp. \bfpage{351}--\blpage{364}
(\byear{2019}).
\bcomment{Springer}
\end{bchapter}
\endbibitem

\bibitem{chang2020spatial}
\begin{barticle}
\bauthor{\bsnm{Chang}, \binits{S.}},
\bauthor{\bsnm{Zhang}, \binits{Y.}},
\bauthor{\bsnm{Zhang}, \binits{F.}},
\bauthor{\bsnm{Zhao}, \binits{X.}},
\bauthor{\bsnm{Huang}, \binits{S.}},
\bauthor{\bsnm{Feng}, \binits{Z.}},
\bauthor{\bsnm{Wei}, \binits{Z.}}:
\batitle{Spatial attention fusion for obstacle detection using mmwave radar and vision sensor}.
\bjtitle{Sensors}
\bvolume{20}(\bissue{4}),
\bfpage{956}
(\byear{2020})
\end{barticle}
\endbibitem

\bibitem{hajri2018real}
\begin{botherref}
\oauthor{\bsnm{Hajri}, \binits{H.}},
\oauthor{\bsnm{Rahal}, \binits{M.-C.}}:
Real time lidar and radar high-level fusion for obstacle detection and tracking with evaluation on a ground truth.
arXiv preprint arXiv:1807.11264
(2018)
\end{botherref}
\endbibitem

\bibitem{gholami2022real}
\begin{barticle}
\bauthor{\bsnm{Gholami}, \binits{F.}},
\bauthor{\bsnm{Khanmirza}, \binits{E.}},
\bauthor{\bsnm{Riahi}, \binits{M.}}:
\batitle{Real-time obstacle detection by stereo vision and ultrasonic data fusion}.
\bjtitle{Measurement}
\bvolume{190},
\bfpage{110718}
(\byear{2022})
\end{barticle}
\endbibitem

\bibitem{bertrand2015bio}
\begin{barticle}
\bauthor{\bsnm{Bertrand}, \binits{O.J.}},
\bauthor{\bsnm{Lindemann}, \binits{J.P.}},
\bauthor{\bsnm{Egelhaaf}, \binits{M.}}:
\batitle{A bio-inspired collision avoidance model based on spatial information derived from motion detectors leads to common routes}.
\bjtitle{PLoS computational biology}
\bvolume{11}(\bissue{11}),
\bfpage{1004339}
(\byear{2015})
\end{barticle}
\endbibitem

\bibitem{kestur2012emulating}
\begin{bchapter}
\bauthor{\bsnm{Kestur}, \binits{S.}},
\bauthor{\bsnm{Park}, \binits{M.S.}},
\bauthor{\bsnm{Sabarad}, \binits{J.}},
\bauthor{\bsnm{Dantara}, \binits{D.}},
\bauthor{\bsnm{Narayanan}, \binits{V.}},
\bauthor{\bsnm{Chen}, \binits{Y.}},
\bauthor{\bsnm{Khosla}, \binits{D.}}:
\bctitle{Emulating mammalian vision on reconfigurable hardware}.
In: \bbtitle{2012 IEEE 20th International Symposium on Field-Programmable Custom Computing Machines},
pp. \bfpage{141}--\blpage{148}
(\byear{2012}).
\bcomment{IEEE}
\end{bchapter}
\endbibitem

\bibitem{poulter2018neurobiology}
\begin{barticle}
\bauthor{\bsnm{Poulter}, \binits{S.}},
\bauthor{\bsnm{Hartley}, \binits{T.}},
\bauthor{\bsnm{Lever}, \binits{C.}}:
\batitle{The neurobiology of mammalian navigation}.
\bjtitle{Current Biology}
\bvolume{28}(\bissue{17}),
\bfpage{1023}--\blpage{1042}
(\byear{2018})
\end{barticle}
\endbibitem

\bibitem{wood2021navigating}
\begin{barticle}
\bauthor{\bsnm{Wood}, \binits{E.R.}},
\bauthor{\bsnm{Dudchenko}, \binits{P.A.}}:
\batitle{Navigating space in the mammalian brain}.
\bjtitle{Science}
\bvolume{372}(\bissue{6545}),
\bfpage{913}--\blpage{914}
(\byear{2021})
\end{barticle}
\endbibitem

\bibitem{de2022predictive}
\begin{barticle}
\bauthor{\bsnm{De~Cothi}, \binits{W.}},
\bauthor{\bsnm{Nyberg}, \binits{N.}},
\bauthor{\bsnm{Griesbauer}, \binits{E.-M.}},
\bauthor{\bsnm{Ghanam{\'e}}, \binits{C.}},
\bauthor{\bsnm{Zisch}, \binits{F.}},
\bauthor{\bsnm{Lefort}, \binits{J.M.}},
\bauthor{\bsnm{Fletcher}, \binits{L.}},
\bauthor{\bsnm{Newton}, \binits{C.}},
\bauthor{\bsnm{Renaudineau}, \binits{S.}},
\bauthor{\bsnm{Bendor}, \binits{D.}}, \betal:
\batitle{Predictive maps in rats and humans for spatial navigation}.
\bjtitle{Current Biology}
\bvolume{32}(\bissue{17}),
\bfpage{3676}--\blpage{3689}
(\byear{2022})
\end{barticle}
\endbibitem

\bibitem{von1990classification}
\begin{barticle}
\bauthor{\bparticle{von~der} \bsnm{Emde}, \binits{G.}},
\bauthor{\bsnm{Schnitzler}, \binits{H.-U.}}:
\batitle{Classification of insects by echolocating greater horseshoe bats}.
\bjtitle{Journal of Comparative Physiology A}
\bvolume{167}(\bissue{3}),
\bfpage{423}--\blpage{430}
(\byear{1990})
\end{barticle}
\endbibitem

\bibitem{moss2018auditory}
\begin{bchapter}
\bauthor{\bsnm{Moss}, \binits{C.F.}}:
\bctitle{Auditory mechanisms of echolocation in bats}.
In: \bbtitle{Oxford Research Encyclopedia of Neuroscience},
(\byear{2018})
\end{bchapter}
\endbibitem

\bibitem{kossl1995basilar}
\begin{barticle}
\bauthor{\bsnm{K{\"o}ssl}, \binits{M.}},
\bauthor{\bsnm{Russell}, \binits{I.}}:
\batitle{Basilar membrane resonance in the cochlea of the mustached bat.}
\bjtitle{Proceedings of the National Academy of Sciences}
\bvolume{92}(\bissue{1}),
\bfpage{276}--\blpage{279}
(\byear{1995})
\end{barticle}
\endbibitem

\bibitem{mead1990neuromorphic}
\begin{barticle}
\bauthor{\bsnm{Mead}, \binits{C.}}:
\batitle{Neuromorphic electronic systems}.
\bjtitle{Proceedings of the IEEE}
\bvolume{78}(\bissue{10}),
\bfpage{1629}--\blpage{1636}
(\byear{1990})
\end{barticle}
\endbibitem

\bibitem{blum2017neuromorphic}
\begin{botherref}
\oauthor{\bsnm{Blum}, \binits{H.}},
\oauthor{\bsnm{Dietm{\"u}ller}, \binits{A.}},
\oauthor{\bsnm{Milde}, \binits{M.}},
\oauthor{\bsnm{Conradt}, \binits{J.}},
\oauthor{\bsnm{Indiveri}, \binits{G.}},
\oauthor{\bsnm{Sandamirskaya}, \binits{Y.}}:
A neuromorphic controller for a robotic vehicle equipped with a dynamic vision sensor.
Robotics Science and Systems, RSS 2017
(2017)
\end{botherref}
\endbibitem

\bibitem{salt2017obstacle}
\begin{bchapter}
\bauthor{\bsnm{Salt}, \binits{L.}},
\bauthor{\bsnm{Indiveri}, \binits{G.}},
\bauthor{\bsnm{Sandamirskaya}, \binits{Y.}}:
\bctitle{Obstacle avoidance with lgmd neuron: towards a neuromorphic uav implementation}.
In: \bbtitle{2017 IEEE International Symposium on Circuits and Systems (ISCAS)},
pp. \bfpage{1}--\blpage{4}
(\byear{2017}).
\bcomment{IEEE}
\end{bchapter}
\endbibitem

\bibitem{hu2016bio}
\begin{barticle}
\bauthor{\bsnm{Hu}, \binits{C.}},
\bauthor{\bsnm{Arvin}, \binits{F.}},
\bauthor{\bsnm{Xiong}, \binits{C.}},
\bauthor{\bsnm{Yue}, \binits{S.}}:
\batitle{Bio-inspired embedded vision system for autonomous micro-robots: The lgmd case}.
\bjtitle{IEEE transactions on cognitive and developmental systems}
\bvolume{9}(\bissue{3}),
\bfpage{241}--\blpage{254}
(\byear{2016})
\end{barticle}
\endbibitem

\bibitem{milde2017obstacle}
\begin{barticle}
\bauthor{\bsnm{Milde}, \binits{M.B.}},
\bauthor{\bsnm{Blum}, \binits{H.}},
\bauthor{\bsnm{Dietm{\"u}ller}, \binits{A.}},
\bauthor{\bsnm{Sumislawska}, \binits{D.}},
\bauthor{\bsnm{Conradt}, \binits{J.}},
\bauthor{\bsnm{Indiveri}, \binits{G.}},
\bauthor{\bsnm{Sandamirskaya}, \binits{Y.}}:
\batitle{Obstacle avoidance and target acquisition for robot navigation using a mixed signal analog/digital neuromorphic processing system}.
\bjtitle{Frontiers in neurorobotics}
\bvolume{11},
\bfpage{28}
(\byear{2017})
\end{barticle}
\endbibitem

\bibitem{dominguez2018deep}
\begin{bchapter}
\bauthor{\bsnm{Dominguez-Morales}, \binits{J.P.}},
\bauthor{\bsnm{Liu}, \binits{Q.}},
\bauthor{\bsnm{James}, \binits{R.}},
\bauthor{\bsnm{Gutierrez-Galan}, \binits{D.}},
\bauthor{\bsnm{Jimenez-Fernandez}, \binits{A.}},
\bauthor{\bsnm{Davidson}, \binits{S.}},
\bauthor{\bsnm{Furber}, \binits{S.}}:
\bctitle{Deep spiking neural network model for time-variant signals classification: a real-time speech recognition approach}.
In: \bbtitle{2018 International Joint Conference on Neural Networks (IJCNN)},
pp. \bfpage{1}--\blpage{8}
(\byear{2018}).
\bcomment{IEEE}
\end{bchapter}
\endbibitem

\bibitem{wall2012spiking}
\begin{barticle}
\bauthor{\bsnm{Wall}, \binits{J.A.}},
\bauthor{\bsnm{McDaid}, \binits{L.J.}},
\bauthor{\bsnm{Maguire}, \binits{L.P.}},
\bauthor{\bsnm{McGinnity}, \binits{T.M.}}:
\batitle{Spiking neural network model of sound localization using the interaural intensity difference}.
\bjtitle{IEEE transactions on neural networks and learning systems}
\bvolume{23}(\bissue{4}),
\bfpage{574}--\blpage{586}
(\byear{2012})
\end{barticle}
\endbibitem

\bibitem{wu2020deep}
\begin{barticle}
\bauthor{\bsnm{Wu}, \binits{J.}},
\bauthor{\bsnm{Y{\i}lmaz}, \binits{E.}},
\bauthor{\bsnm{Zhang}, \binits{M.}},
\bauthor{\bsnm{Li}, \binits{H.}},
\bauthor{\bsnm{Tan}, \binits{K.C.}}:
\batitle{Deep spiking neural networks for large vocabulary automatic speech recognition}.
\bjtitle{Frontiers in neuroscience}
\bvolume{14},
\bfpage{199}
(\byear{2020})
\end{barticle}
\endbibitem

\bibitem{arena2009learning}
\begin{barticle}
\bauthor{\bsnm{Arena}, \binits{P.}},
\bauthor{\bsnm{Fortuna}, \binits{L.}},
\bauthor{\bsnm{Frasca}, \binits{M.}},
\bauthor{\bsnm{Patan{\'e}}, \binits{L.}}:
\batitle{Learning anticipation via spiking networks: application to navigation control}.
\bjtitle{IEEE transactions on neural networks}
\bvolume{20}(\bissue{2}),
\bfpage{202}--\blpage{216}
(\byear{2009})
\end{barticle}
\endbibitem

\bibitem{furber2014spinnaker}
\begin{barticle}
\bauthor{\bsnm{Furber}, \binits{S.B.}},
\bauthor{\bsnm{Galluppi}, \binits{F.}},
\bauthor{\bsnm{Temple}, \binits{S.}},
\bauthor{\bsnm{Plana}, \binits{L.A.}}:
\batitle{{The SpiNNaker project}}.
\bjtitle{Proceedings of the IEEE}
\bvolume{102}(\bissue{5}),
\bfpage{652}--\blpage{665}
(\byear{2014})
\end{barticle}
\endbibitem

\bibitem{rowley2019spinntools}
\begin{barticle}
\bauthor{\bsnm{Rowley}, \binits{A.G.}},
\bauthor{\bsnm{Brenninkmeijer}, \binits{C.}},
\bauthor{\bsnm{Davidson}, \binits{S.}},
\bauthor{\bsnm{Fellows}, \binits{D.}},
\bauthor{\bsnm{Gait}, \binits{A.}},
\bauthor{\bsnm{Lester}, \binits{D.R.}},
\bauthor{\bsnm{Plana}, \binits{L.A.}},
\bauthor{\bsnm{Rhodes}, \binits{O.}},
\bauthor{\bsnm{Stokes}, \binits{A.B.}},
\bauthor{\bsnm{Furber}, \binits{S.B.}}:
\batitle{Spinntools: the execution engine for the spinnaker platform}.
\bjtitle{Frontiers in neuroscience}
\bvolume{13},
\bfpage{231}
(\byear{2019})
\end{barticle}
\endbibitem

\bibitem{ghosh2009spiking}
\begin{barticle}
\bauthor{\bsnm{Ghosh-Dastidar}, \binits{S.}},
\bauthor{\bsnm{Adeli}, \binits{H.}}:
\batitle{Spiking neural networks}.
\bjtitle{International journal of neural systems}
\bvolume{19}(\bissue{04}),
\bfpage{295}--\blpage{308}
(\byear{2009})
\end{barticle}
\endbibitem

\bibitem{davidson2021comparison}
\begin{barticle}
\bauthor{\bsnm{Davidson}, \binits{S.}},
\bauthor{\bsnm{Furber}, \binits{S.B.}}:
\batitle{Comparison of artificial and spiking neural networks on digital hardware}.
\bjtitle{Frontiers in Neuroscience}
\bvolume{15},
\bfpage{345}
(\byear{2021})
\end{barticle}
\endbibitem

\bibitem{lobo2020spiking}
\begin{barticle}
\bauthor{\bsnm{Lobo}, \binits{J.L.}},
\bauthor{\bsnm{Del~Ser}, \binits{J.}},
\bauthor{\bsnm{Bifet}, \binits{A.}},
\bauthor{\bsnm{Kasabov}, \binits{N.}}:
\batitle{Spiking neural networks and online learning: An overview and perspectives}.
\bjtitle{Neural Networks}
\bvolume{121},
\bfpage{88}--\blpage{100}
(\byear{2020})
\end{barticle}
\endbibitem

\bibitem{davison2009pynn}
\begin{barticle}
\bauthor{\bsnm{Davison}, \binits{A.P.}},
\bauthor{\bsnm{Br{\"u}derle}, \binits{D.}},
\bauthor{\bsnm{Eppler}, \binits{J.M.}},
\bauthor{\bsnm{Kremkow}, \binits{J.}},
\bauthor{\bsnm{Muller}, \binits{E.}},
\bauthor{\bsnm{Pecevski}, \binits{D.}},
\bauthor{\bsnm{Perrinet}, \binits{L.}},
\bauthor{\bsnm{Yger}, \binits{P.}}:
\batitle{Pynn: a common interface for neuronal network simulators}.
\bjtitle{Frontiers in neuroinformatics}
\bvolume{2},
\bfpage{11}
(\byear{2009})
\end{barticle}
\endbibitem

\bibitem{rhodes2018spynnaker}
\begin{botherref}
\oauthor{\bsnm{Rhodes}, \binits{O.}},
\oauthor{\bsnm{Bogdan}, \binits{P.A.}},
\oauthor{\bsnm{Brenninkmeijer}, \binits{C.}},
\oauthor{\bsnm{Davidson}, \binits{S.}},
\oauthor{\bsnm{Fellows}, \binits{D.}},
\oauthor{\bsnm{Gait}, \binits{A.}},
\oauthor{\bsnm{Lester}, \binits{D.R.}},
\oauthor{\bsnm{Mikaitis}, \binits{M.}},
\oauthor{\bsnm{Plana}, \binits{L.A.}},
\oauthor{\bsnm{Rowley}, \binits{A.G.}}, et al.:
spynnaker: a software package for running pynn simulations on spinnaker.
Frontiers in neuroscience,
816
(2018)
\end{botherref}
\endbibitem

\bibitem{hines1997neuron}
\begin{barticle}
\bauthor{\bsnm{Hines}, \binits{M.L.}},
\bauthor{\bsnm{Carnevale}, \binits{N.T.}}:
\batitle{The neuron simulation environment}.
\bjtitle{Neural computation}
\bvolume{9}(\bissue{6}),
\bfpage{1179}--\blpage{1209}
(\byear{1997})
\end{barticle}
\endbibitem

\bibitem{Gewaltig:NEST}
\begin{barticle}
\bauthor{\bsnm{Gewaltig}, \binits{M.-O.}},
\bauthor{\bsnm{Diesmann}, \binits{M.}}:
\batitle{Nest (neural simulation tool)}.
\bjtitle{Scholarpedia}
\bvolume{2}(\bissue{4}),
\bfpage{1430}
(\byear{2007})
\end{barticle}
\endbibitem

\bibitem{goodman2008brian}
\begin{barticle}
\bauthor{\bsnm{Goodman}, \binits{D.F.}},
\bauthor{\bsnm{Brette}, \binits{R.}}:
\batitle{Brian: a simulator for spiking neural networks in python}.
\bjtitle{Frontiers in neuroinformatics}
\bvolume{2},
\bfpage{5}
(\byear{2008})
\end{barticle}
\endbibitem

\bibitem{auge2021survey}
\begin{barticle}
\bauthor{\bsnm{Auge}, \binits{D.}},
\bauthor{\bsnm{Hille}, \binits{J.}},
\bauthor{\bsnm{Mueller}, \binits{E.}},
\bauthor{\bsnm{Knoll}, \binits{A.}}:
\batitle{A survey of encoding techniques for signal processing in spiking neural networks}.
\bjtitle{Neural Processing Letters}
\bvolume{53}(\bissue{6}),
\bfpage{4693}--\blpage{4710}
(\byear{2021})
\end{barticle}
\endbibitem

\end{thebibliography}

\end{document}